\documentclass[letterpaper, 10 pt, conference]{ieeeconf}  

\IEEEoverridecommandlockouts                              

\usepackage{graphics} 
\usepackage{epsfig} 
\usepackage{mathptmx} 
\usepackage{times} 
\usepackage[OT1]{fontenc} 
\usepackage{amsmath,amssymb,amsfonts}  
\usepackage{multirow}
\usepackage{multicol}
\usepackage{cite}
\usepackage{algorithmic}
\usepackage{textcomp}
\usepackage{xcolor}
\usepackage{url}

\title{\LARGE \bf
Open-Source Autonomous Driving Software Platforms: \\ Comparison of Autoware and Apollo
}

\author{Hee-Yang Jung\textsuperscript{1}, Dong-Hee Paek\textsuperscript{2} and Seung-Hyun Kong\textsuperscript{*}
\thanks{*corresponding author}
\thanks{\textsuperscript{1}Hee-Yang Jung is with the Robotics Program, Korea Advanced Institute of Science and Technology, Daejeon, Korea, 34051
        {\tt\small heeyang@kaist.ac.kr}}%
\thanks{\textsuperscript{2}Dong-Hee Paek and \textsuperscript{*}Seung-Hyun Kong are with the CCS Graduate School of Mobility, Korea Advanced Institute of Science and Technology, Daejeon, Korea, 34051
        {\tt\small \{donghee.paek,skong\} @kaist.ac.kr}}
}

\begin{document}

\maketitle
\thispagestyle{empty}
\pagestyle{empty}

\begin{abstract}

Full-stack autonomous driving system spans diverse technological domains—including perception, planning, and control—that each require in-depth research. Moreover, validating such technologies of the system necessitates extensive supporting infrastructure, from simulators and sensors to high-definition maps. These complexities with barrier to entry pose substantial limitations for individual developers and research groups. Recently, open-source autonomous driving software platforms have emerged to address this challenge by providing autonomous driving technologies and practical supporting infrastructure for implementing and evaluating autonomous driving functionalities. Among the prominent open-source platforms, Autoware and Apollo are frequently adopted in both academia and industry. While previous studies have assessed each platform independently, few have offered a quantitative and detailed head-to-head comparison of their capabilities. In this paper, we systematically examine the core modules of Autoware and Apollo and evaluate their middleware performance to highlight key differences. These insights serve as a practical reference for researchers and engineers, guiding them in selecting the most suitable platform for their specific development environments and advancing the field of full-stack autonomous driving system.

\end{abstract}

\section{Introduction}
A full-stack autonomous driving system perceives its surrounding environment by processing data collected from various sensors and then plans the vehicle’s future actions accordingly. The computed plan is subsequently transmitted through a controller to the vehicle’s throttle, brake, and steering systems. Developing such a system requires a comprehensive understanding of sensor technology, perception and path planning algorithms, and effective vehicle control techniques—each of which must be seamlessly integrated. In addition to this complexity, implementing autonomous driving technology demands an extensive infrastructure, including simulators for validation and appropriate sensors for real-world testing. These challenges impose substantial limitations on individual developers or research groups attempting to independently build a full-stack autonomous driving system \cite{kato2015open, raju2019performance}.

\begin{figure}[t]
    \centerline{\includegraphics[width=0.9\columnwidth]{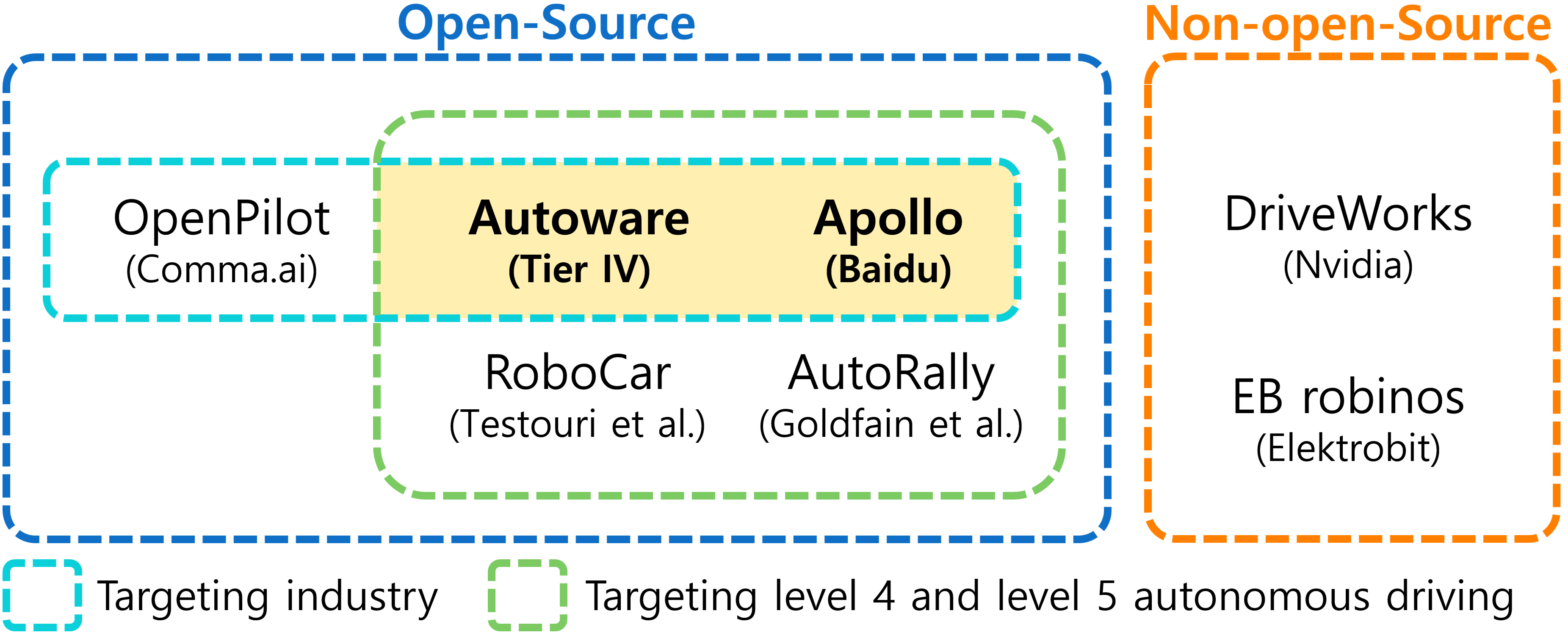}}
    \caption{
        Classification of autonomous driving software platforms.
    }
    \label{software platform classification}
\end{figure}

Since the 2010s, numerous vendors and research organizations have actively developed and distributed autonomous driving software platforms to lower this barrier. Examples from industry include NVIDIA’s DriveWorks \cite{nvidia2018driveworks}, Elektrobit’s EB robinos \cite{elektrobit}, Tier IV’s Autoware \cite{autoawre_github}, and Baidu’s Apollo \cite{apollo_github}. Additionally, comma.ai introduced OpenPilot \cite{commaai_openpilot}, which offers Advanced Driver Assistance System (ADAS) functionalities. Beyond these industry efforts, research organizations have contributed platforms like RoboCar \cite{testouri2024robocarrapidlydeployableopensource} and AutoRally \cite{goldfain2019autorally}. Depending on factors such as openness, target application, and autonomy level, autonomous driving software platforms can be classified into distinct categories (Fig. \ref{software platform classification}). For instance, non-open-source platforms, e.g., DriveWorks and EB robinos, limit modifications and redistribution, potentially hindering the advancement of autonomous driving technology \cite{bulwahn2013research}. Conversely, open-source platforms, e.g., OpenPilot, Autoware, Apollo, RoboCar, and AutoRally, benefit from global community-driven maintenance and can evolve more rapidly and flexibly, thereby increasing the likelihood of standardization \cite{wright2014open}.

Among open-source platforms, scalability and autonomy level are key considerations in full-stack autonomous driving system. Platforms designed for industrial applications, such as OpenPilot, Autoware, and Apollo, are built with scalability in mind and can be integrated into various vehicles. In contrast, those tailored to specific vehicle models, such as RoboCar and AutoRally, are primarily used for research and experimentation, making broad applicability more challenging. Meanwhile, achieving complete autonomous driving requires platforms that support at least level 4 autonomy. OpenPilot, for example, focuses mainly on ADAS functionalities, rendering it unsuitable for higher-level autonomy \cite{alsubaei2022reliability}. In contrast, both Autoware and Apollo offer Level 4 or higher autonomy technologies to enable complete autonomous driving. Hence, commercializing autonomous driving calls for open-source platforms that are both scalable to industrial applications and capable of achieving level 4 or higher autonomy—criteria well satisfied by Autoware and Apollo.

\begin{figure*}[tb]
    \centerline{\includegraphics[width=0.9\linewidth]{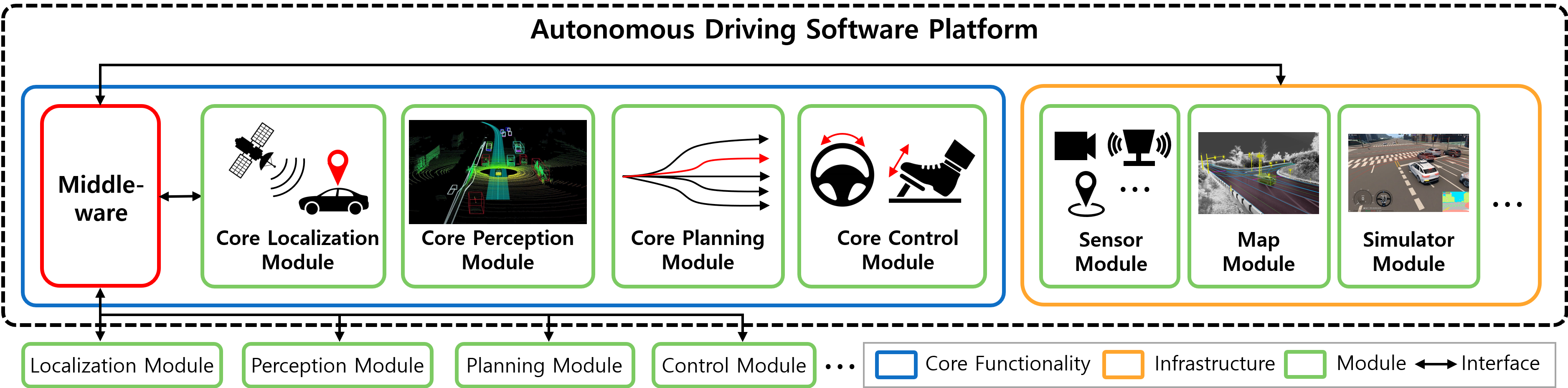}}
    \caption{
        Architecture of autonomous driving software platforms. Core functionality includes middleware for data communication between modules. The core modules perform essential functions required for full-stack autonomous driving, such as localization, perception, planning, and control. The infrastructure consists of sensors, maps, and simulators necessary for operating these modules.
    }
    \label{autonomous driving software platform architecture}
\end{figure*} 

\begin{figure}[t]
    \centerline{\includegraphics[width=1.0\columnwidth]{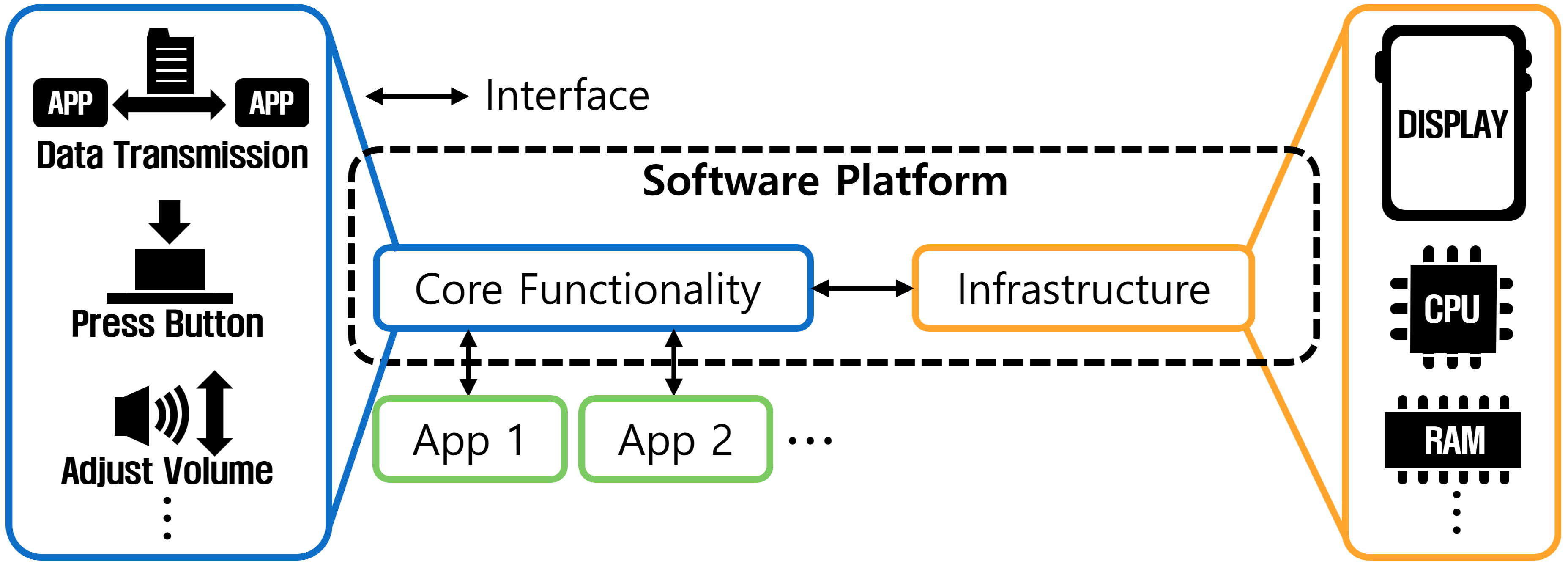}}
    \caption{
        Architecture of software platforms.
    }
    \label{software platform architecture}
\end{figure}
Previous survey studies have largely focused on Autoware or Apollo in isolation. For instance, \cite{kato2018autoware} provides an overview of Autoware’s software stack and its functionalities, and \cite{bulwahn2013research} validates the feasibility of applying Autoware to real vehicles. Meanwhile, \cite{fan2018baidu} delves into Apollo’s optimal path planning techniques, and \cite{feng2022application} proposes a novel control approach based on Apollo. Although other works mention both platforms, few provide an in-depth, direct comparison. For example, \cite{raju2019performance} qualitatively discusses the overall architectures of Autoware and Apollo, while \cite{garcia2020comprehensive} analyzes bugs from both platforms. In \cite{andreigavrilov2021analysis}, the focus is on data communication performance, and \cite{wang2024moving} categorizes autonomous driving approaches to highlight modular versus end-to-end strategies. these studies do not offer a systematic, quantitative comparison of how each platform implements its core functionalities. Consequently, researchers and engineers lack clear guidance on selecting the most appropriate open-source autonomous driving software platform for their specific projects.

To address this gap, this paper presents a comparative analysis of Autoware and Apollo, focusing on both their core functionalities and middleware performance. By examining each system’s localization, perception, planning, and control modules. We illuminate key differences that matter from a development standpoint. The remainder of this paper is organized as follows. Section II defines the concept of software platforms and autonomous driving software platforms. Section III provides a comparative analysis of Autoware and Apollo, detailing their underlying architectures and highlighting each platform’s strengths and limitations. In Section IV, we quantify the functionalities offered by both platforms and evaluate middleware performance to assess how effectively each handles large-scale data transmission with minimal latency and error.

\begin{figure}[t]
    \centerline{\includegraphics[width=1.0\columnwidth]{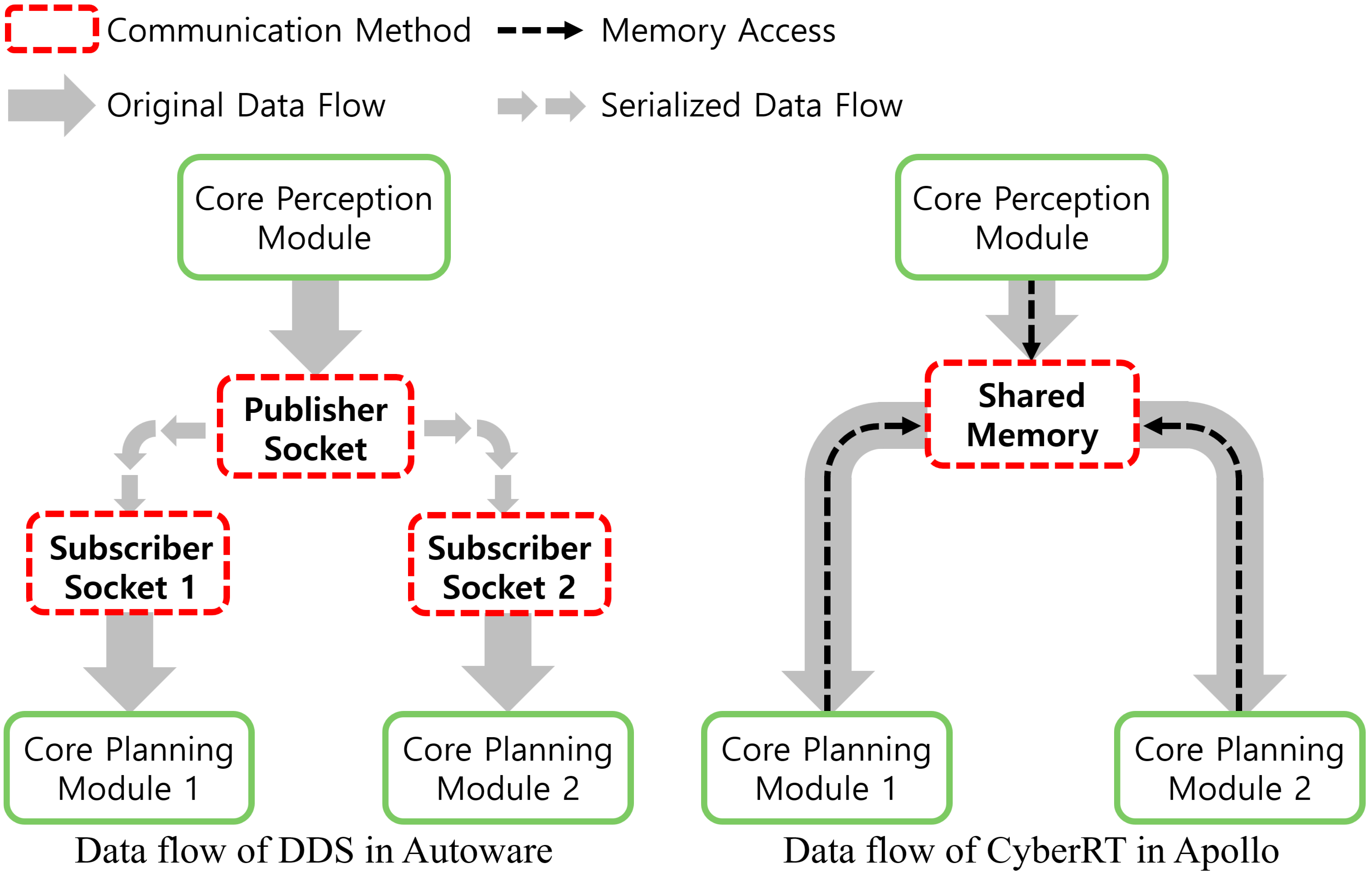}}
    \caption{
        Data flow of the middleware used in Autoware and Apollo. Raw data is transmitted from the core perception module. In Autoware, data is fragmented through serialization between the publisher socket and the subscriber socket. In contrast, Apollo transmits raw data directly to shared memory without serialization, allowing the core planning modules to access the raw data directly from shared memory.
    }
    \label{The data flow in middleware}
\end{figure}


\section{Autonomous Driving Software Platform}
To compare Autoware and Apollo, we first clarify what constitutes an autonomous driving software platform. This concept builds on the general definition of a software platform, which Tiwana et al. describe as \textit{“a foundation on which outside parties can build complementary products or services, providing the core functionality shared by apps”} \cite{tiwana2010research, tiwana2013platform}. In essence, a software platform provides core functionality and the necessary infrastructure, enabling developers to create various applications.

A typical software platform architecture is depicted in Fig. \ref{software platform architecture}. The \textit{core functionality}  offers fundamental features required for application development, e.g., data transmission, button interactions, and volume control. The \textit{infrastructure} comprises resources—such as displays, CPUs, and RAM—supporting the operation of these features. The \textit{interface} connects the core functionality with both the applications and the infrastructure, enabling seamless data communication among them.

Extending this concept to autonomous driving, an autonomous driving software platform integrates additional requirements to meet the demands of real-world deployment. Fig. \ref{autonomous driving software platform architecture} outlines its architecture. The core functionality comprises (1) middleware for data communication and (2) core modules responsible for the end-to-end autonomous driving pipeline. Unlike generic software platforms, autonomous driving platforms must handle real-time data from diverse sensors and manage computationally intensive tasks, e.g., perception and planning. Consequently, robust middleware and cohesive core modules are essential for functionality such as localization, perception, planning, and control.
The infrastructure in an autonomous driving context includes sensor modules, map databases, simulators, and other critical resources required to test, validate, and deploy autonomous driving algorithms. The interface ensures seamless data transmission among modules and the underlying infrastructure, thereby enabling stable and efficient operation of the entire system.

\begin{figure}[t]
    \centerline{\includegraphics[width=1.0\columnwidth]{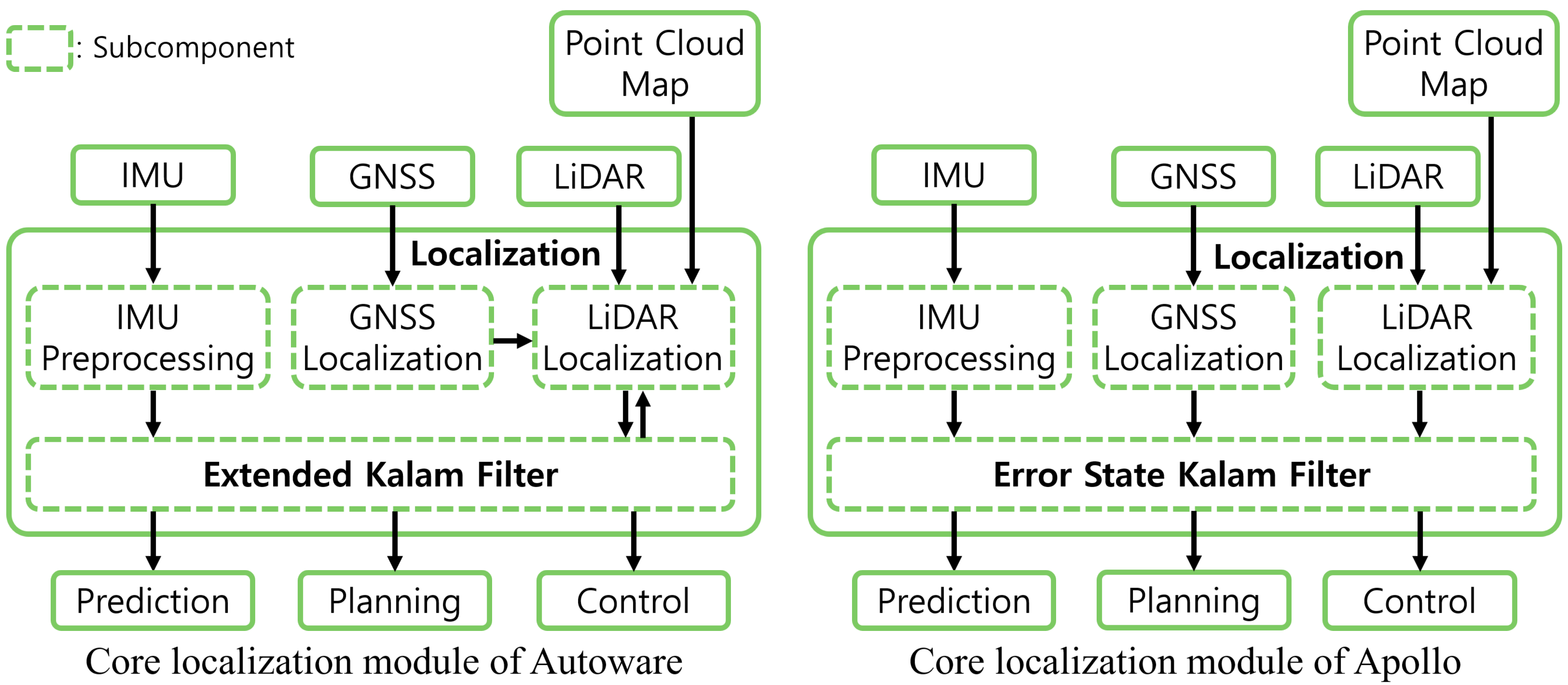}}
    \caption{
        Internal structure of the core localization module used in Autoware and Apollo. The module estimates position using IMU, GNSS, and LiDAR. In LiDAR localization, a pre-built point cloud map is utilized to determine the vehicle's position. The position estimates obtained from each sensor are integrated using the Kalman Filter technique to enhance accuracy.
    }
    \label{core localization module}
\end{figure} 


\section{Open-Source Autonomous Driving Software Platform Comparison}

Having defined the general architecture of an autonomous driving software platform, we now examine how this applies to Autoware \cite{autoawre_github} and Apollo \cite{apollo_github}. In particular, we analyze (1) middleware for data communication and (2) the core modules that execute the sequential tasks of a full-stack autonomous driving system. As illustrated in Fig. \ref{autonomous driving software platform architecture}, a full-stack pipeline typically proceeds from localization to perception, then planning, and finally control.

\subsection{Middleware}
Middleware is a core component responsible for data transmission between modules (Fig. \ref{The data flow in middleware}). Autoware is based on ROS2, which employs Data Distribution Service (DDS) as middleware \cite{maruyama2016exploring}. DDS creates TCP or UDP sockets between a publisher and a subscriber and serializes large-scale data before transmission \cite{liu2020memory}. This involves fragmenting data into multiple packets, which are reassembled upon reception. This approach can increase bottlenecks as data size grows.
In contrast, Apollo uses a proprietary middleware called CyberRT. CyberRT allocates a shared memory space, allowing the publisher to write data directly and the subscriber to read it \cite{liu2020robotic}. This design obviates the need for serialization and can significantly improve data transmission speed. However, if simultaneous read/write operations occur without proper synchronization, data corruption may arise. Consequently, Apollo’s approach requires careful management to prevent race conditions.

\begin{figure}[t]
    \centerline{\includegraphics[width=1.0\columnwidth]{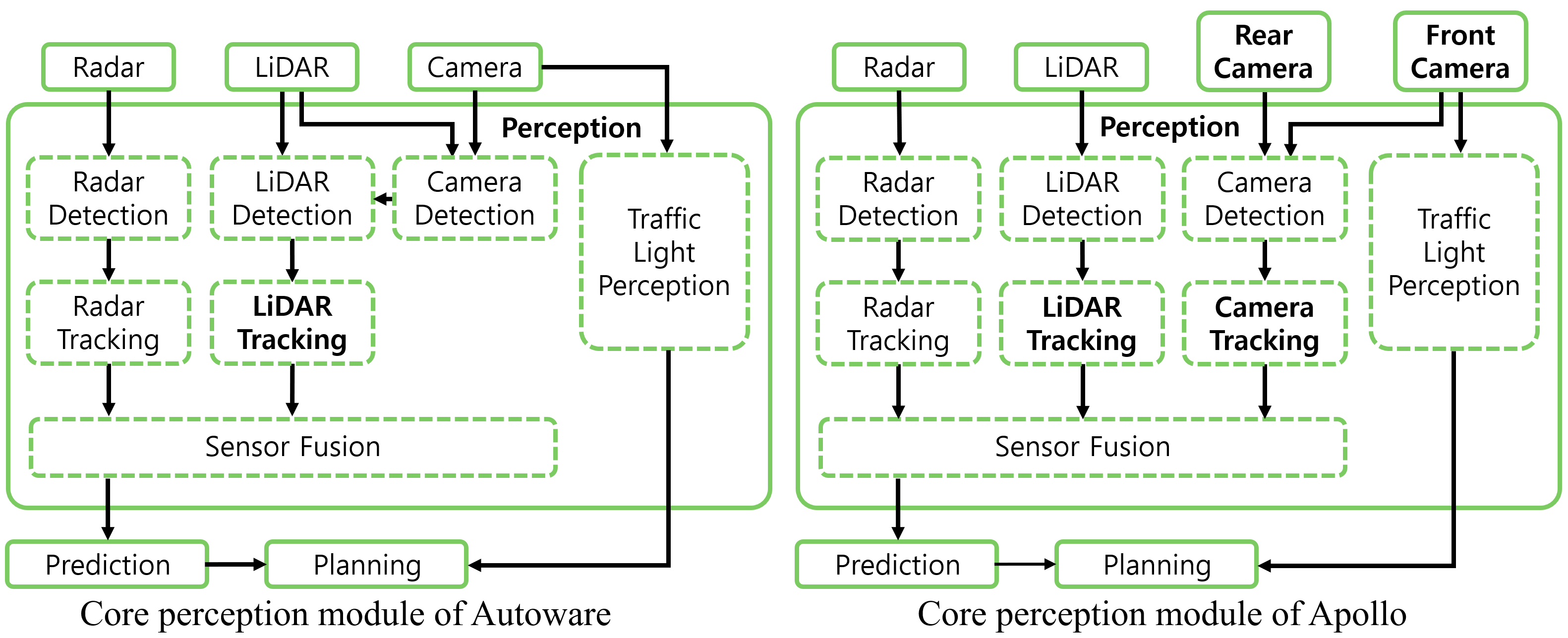}}
    \caption{
        Internal structure of the core perception module used in Autoware and Apollo. Data is acquired from vision sensors such as radar, LiDAR, and camera. The data is processed through detection algorithms designed for each sensor and then transmitted to the prediction and perception modules.
    }
    \label{core perception module}
\end{figure}

\subsection{Localizaiton}
Localization is the starting point of the autonomous driving pipeline, tasked with estimating the ego-vehicle’s position using sensors such as IMU, GNSS, LiDAR, and cameras. Fig. \ref{core localization module} illustrates how sensor data is combined to determine the ego-vehicle’s pose.
Both Autoware and Apollo employ IMU and GNSS as primary localization sensors, using similar noise-filtering techniques for the IMU data but diverging in their GNSS localization strategies \cite{autoware_universe_docs, wan2018robust}. While Autoware estimates position based solely on satellite signals, Apollo refines GNSS-based localization by incorporating correction data from reference stations such as Real-Time Kinematic (RTK) \cite{wan2018robust}.
LiDAR-based localization is another critical component. Autoware uses the Normal Distributions Transform (NDT) \cite{1249285}, initializing position with covariance estimates from an Extended Kalman Filter (EKF). Apollo, on the other hand, projects LiDAR point clouds onto a Bird’s-Eye View (BEV) and applies the Lucas–Kanade \cite{lucas1981iterative} method combined with a histogram filter \cite{thrun2002probabilistic} to reduce computational complexity. 
Both platforms fuse multiple sensor outputs with Kalman Filter-based methods—Autoware via an EKF \cite{ribeiro2004kalman} and Apollo via an Error State Kalman Filter (ESKF) \cite{sola2017quaternion}—to enhance accuracy, while Apollo’s ESKF also reduces linearization errors for more stable convergence.

\subsection{Perception}
The perception module acquires data from radar, LiDAR, and camera sensors to identify the surrounding environment, extracting object attributes such as position, size, and orientation (Fig. \ref{core perception module}). After processing these sensor inputs, the module transmits the resulting information to the prediction and planning modules.
Radar is typically used for long-range detection. Autoware employs model-based methods, whereas Apollo supports not only model-based but also data-driven approaches like PointPillars \cite{lang2019pointpillars} for increased robustness.
LiDAR provides detailed three-dimensional information, and camera images enrich semantic understanding, e.g., color, shape, and texture. Autoware enhances 3D object detection by combining model-based approaches such as L-Shape Fitting \cite{zhang2017efficient} with data-driven techniques such as PointPillars, YOLOX \cite{DBLP:journals/corr/abs-2107-08430}, and Frustum PointNets \cite{qi2018frustum}. On the other hand, Apollo similarly offers data-driven techniques like CenterPoint \cite{yin2021center}, PointPillars, CNNSeg, and MaskPillars, and it utilizes both front and rear cameras to minimize collision risks.
Both platforms also handle traffic light detection via camera inputs, analyzing images to determine traffic signal states, which are subsequently passed to the planning module.

\begin{figure}[t]
    \centerline{\includegraphics[width=1.0\columnwidth]{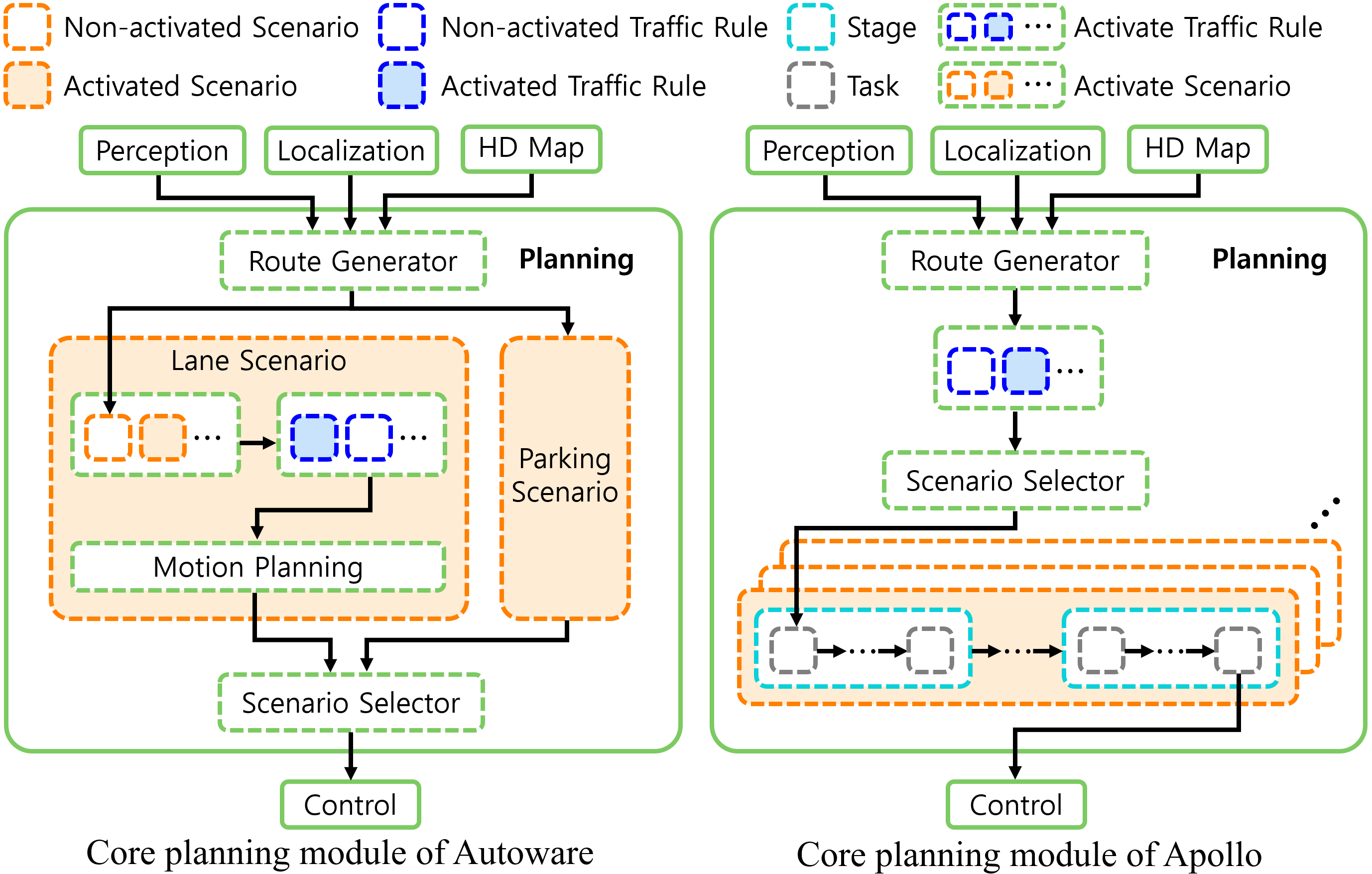}}
    \caption{
        Internal structure of the core planning module used in Autoware and Apollo. Only a subset of scenarios and traffic rules are activated. The activated scenario generates a trajectory, which is then transmitted to the control module.
    }
    \label{core planning module}
\end{figure} 

\subsection{Planning}
The planning module generates a trajectory for future maneuvers using the ego-vehicle’s position and information of surrounding objects. Fig. \ref{core planning module} compares the architecture for the planning module in Autoware and Apollo.
Autoware’s planning process starts by generating a route, followed by parallel execution of lane and parking scenarios. A scenario selector then activates one scenario based on the vehicle’s proximity to its destination. Within the lane scenario, subcomponents handle driving maneuvers, e.g., lane keeping, lane changing, intersection management, and traffic signal handling, and enforce relevant traffic rules. Finally, a collision-free trajectory is planned in the motion planning.
Apollo, by contrast, integrates traffic regulations and environmental factors before choosing a single scenario via a scenario selector. The selected scenario runs multiple sequential stages, each responsible for specific tasks. This architecture can reduce computational overhead compared to Autoware, as Apollo does not compute trajectories for inactive scenarios.

\subsection{Control}
\begin{figure}[t]
    \centerline{\includegraphics[width=0.9\columnwidth]{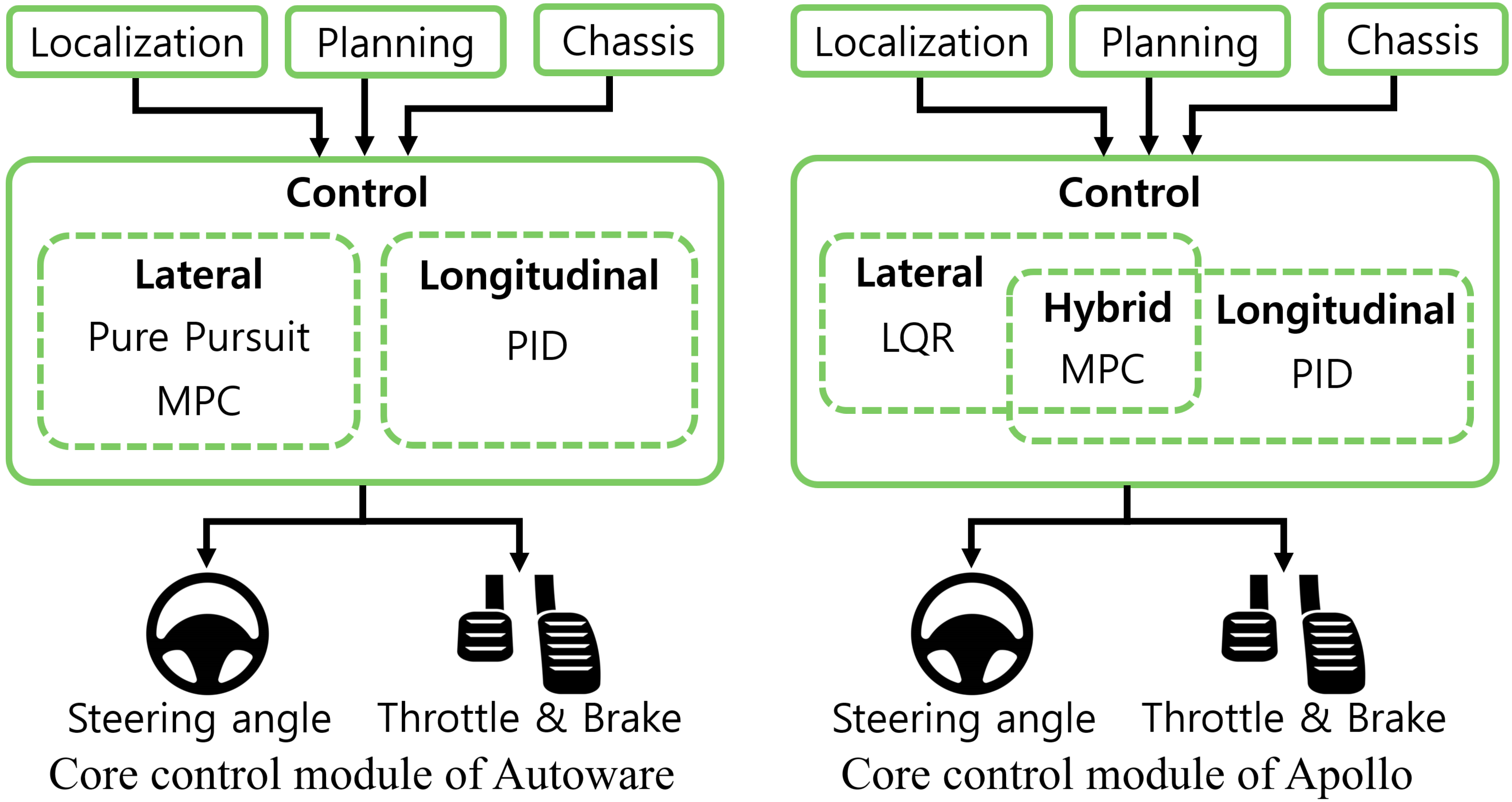}}
    \caption{
        Internal structure of the core control module used in Autoware and Apollo. Autoware provides separate lateral and longitudinal controllers, whereas Apollo also offers a hybrid controller that integrates both.
    }
    \label{Core control module}
\end{figure}

The control module ensures the vehicle follows the generated trajectory by managing lateral (steering) and longitudinal (throttle/brake) control. Fig. \ref{Core control module} summarizes the controllers each platform offers.
Autoware provides two lateral control options—Pure Pursuit and Model Predictive Control (MPC)—while employing a PID controller for longitudinal control. Apollo, on the other hand, uses a Linear Quadratic Regulator (LQR) for lateral control and a PID controller for longitudinal control. Additionally, Apollo offers a hybrid MPC-based controller that integrates lateral and longitudinal control, often delivering enhanced performance when the two controls strongly interact.

\section{Experiments between Autoware and Apollo}
This section presents a quantitative analysis of Autoware and Apollo by examining: (1) the subcomponents in each core module and (2) the performance of the middleware responsible for data transmission. We first describe our method for identifying and comparing the the subcomponents in localization, perception, planning, and control. We then detail our experimental setup for middleware testing, followed by a discussion of the results.

\subsection{Comparison of Core Module Subcomponents}
To investigate functional coverage, we conducted a detailed review of the source code \cite{autoawre_github, autoware_universe_github, apollo_github} and official documentation \cite{autoware_docs, autoware_universe_docs, apollo_docs} for both Autoware and Apollo. We focused on the four primary modules essential for full-stack autonomous driving—localization, perception, planning, and control—and enumerated their respective subcomponents.

\begin{figure*}[t!]
    \centerline{\includegraphics[width=1.0\linewidth]{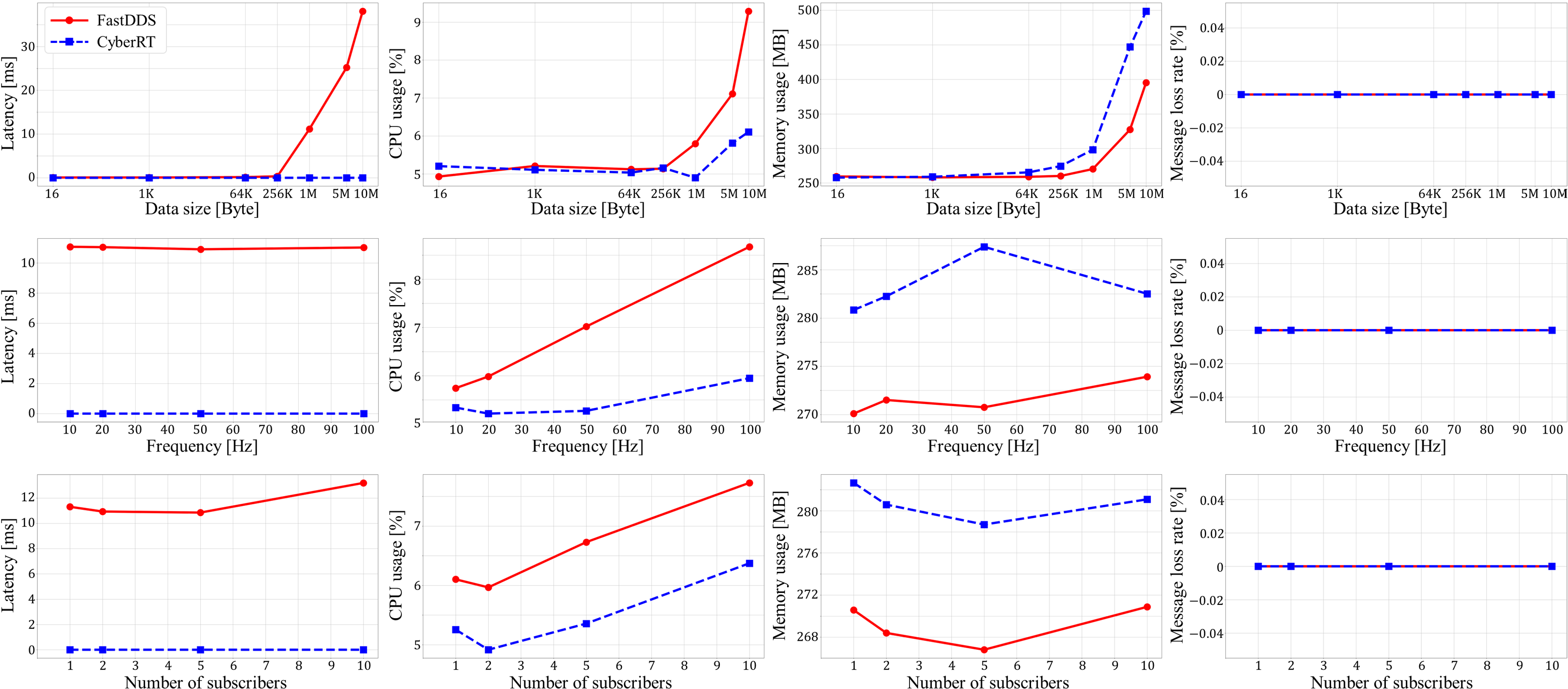}}
    \caption{
        Middleware communication performance results based on data size, frequency, and number of subscribers. The first row represents data size, the second row represents frequency, and the third row represents the number of subscribers. The first column indicates latency, the second column indicates CPU usage, the third column indicates memory usage, and the fourth column indicates message loss rate.
    }
    \label{middleware performance evaluation}
\end{figure*} 

Table \ref{tab:comparison_num_of_functions} summarizes the number of subcomponents in each module. The results indicate that Autoware offers a broader range of localization options, including camera-based and data-driven methods, whereas Apollo provides extensive support for camera-based approaches in perception. For planning, the two platforms differ in the number of available scenarios and traffic regulations. Although the control modules have the same number of sub-functions, the controller designs vary, reflecting each platform’s distinct architectural choices.

\subsection{Middleware Performance Evaluation}
The purpose of middleware is to enable the rapid and error-free transmission of large-scale data between modules in autonomous driving software. Therefore, key performance metrics include transmission latency, CPU usage, memory consumption, and message loss rate.
This paper compares the middleware used in each platform: DDS, i.e., FastDDS\cite{eprosima_fastdds, bode2023systematic}, in Autoware and CyberRT in Apollo. We ran our experiments on a system with an AMD Ryzen 7 5800X 8-Core Processor (16 logical cores, 2.2 GHz), 64 GB of RAM, and a TCP configuration for FastDDS to minimize message loss. In addition, we examined how data transmission performance varies according to (1) module frequency requirements, e.g., typical functional modules versus high-frequency modules, and (2) sensor data size, e.g., standard sensor data versus high-end sensor data.

\begin{table}[t]
    \caption{Number of Subcomponents in Autonomous Driving Software Platforms}
    \label{tab:comparison_num_of_functions}
    \centering
    \renewcommand{\arraystretch}{1.1}
    \setlength{\tabcolsep}{1pt}
    \begin{tabular}{c|c|ccc|ccc|c}
    \hline \hline

    \multirow{2}{*}{} & \multirow{3}{*}{\begin{tabular}[c]{@{}c@{}}Locali-\\ zaiton\end{tabular}} & \multicolumn{3}{c|}{Perception} & \multicolumn{3}{c|}{Planning} & \multirow{2}{*}{Control} \\ \cline{3-8} & & \multicolumn{1}{c}{Radar} & \multicolumn{1}{c}{LiDAR} & Camera & \multicolumn{1}{c}{Scenario} & \multicolumn{1}{c}{\begin{tabular}[c]{@{}c@{}}Traffic\\ rule\end{tabular}} & Task & 
    
    \\ \hline Autoware & \textbf{5} & \multicolumn{1}{c}{\textbf{5}} & \multicolumn{1}{c}{\textbf{17}} & 2 & \multicolumn{1}{c}{7} & \multicolumn{1}{c}{\textbf{14}} & - & 3 
    
    \\ \hline Apollo & 3 & \multicolumn{1}{c}{2} & \multicolumn{1}{c}{9} & \textbf{13} & \multicolumn{1}{c}{\textbf{12}} & \multicolumn{1}{c}{9} & \textbf{21} & 3 
    
    \\ \hline \hline
\end{tabular}
\end{table}

\begin{table}[h]
    \caption{Data Transmission Performances}
    \label{tab:comparison_performance_of_middleware}
    \centering
    \renewcommand{\arraystretch}{1.3}
    \setlength{\tabcolsep}{1pt}
    \begin{tabular}{c|c|cccc}
    \hline \hline

    Modules & Platforms & \begin{tabular}[c]{@{}c@{}}Latency\\ {[}$\mu$s{]}\end{tabular} & \begin{tabular}[c]{@{}c@{}}CPU \\ Usage {[}\%{]}\end{tabular} & \begin{tabular}[c]{@{}c@{}}Memory \\ usage {[}MB{]}\end{tabular} & \begin{tabular}[c]{@{}c@{}}Message \\ loss rate {[}\%{]}\end{tabular} \\ \hline

    \multirow{2}{*}{\begin{tabular}[c]{@{}c@{}}Functional\\module\end{tabular}}
        & FastDDS   & 153           & 5.12          & \textbf{258.89} & 0.0 \\ 
        & CyberRT   & \textbf{0.11} & \textbf{5.03} & 265.39          & 0.0 \\ \hline
    \multirow{2}{*}{\begin{tabular}[c]{@{}c@{}}High frequency \\module\end{tabular}}
        & FastDDS   & 139           & 5.93          & \textbf{259.67} & 0.0 \\  
        & CyberRT   & \textbf{0.05} & \textbf{5.29} & 262.46          & 0.0 \\ \hline
    \multirow{2}{*}{\begin{tabular}[c]{@{}c@{}}Normal sensor \\module\end{tabular}}
        & FastDDS   & 11126         & 5.79          & \textbf{270.19} & 0.0 \\ 
        & CyberRT   & \textbf{0.27} & \textbf{4.90} & 298.18          & 0.0 \\ \hline
    \multirow{2}{*}{\begin{tabular}[c]{@{}c@{}}High-end sensor\\module\end{tabular}}
        & FastDDS   & 38075         & 9.29          & \textbf{395.30} & 0.0 \\ 
        & CyberRT   & \textbf{1.72} & \textbf{6.11} & 498.72          & 0.0 \\ \hline \hline

\end{tabular}
\end{table}
Fig. \ref{middleware performance evaluation} provides a visual comparison of how FastDDS and CyberRT perform under varying data sizes, frequencies, and numbers of subscribers, while Table \ref{tab:comparison_performance_of_middleware} summarizes the numeric results.
FastDDS exhibits a significant increase in latency when data sizes grow, primarily due to overhead from multiple serialization processes, which in turn leads to higher CPU usage. By contrast, CyberRT’s shared memory approach allows consistently lower latency across different workloads, reducing the need for serialization.
However, CyberRT’s reliance on shared memory allocation results in higher overall memory usage—tens of megabytes more than FastDDS in Table \ref{tab:comparison_performance_of_middleware}. Despite these differences in design trade-offs, neither middleware exhibited message loss under the tested conditions. CPU usage was relatively similar between the two, although slight variations were observed depending on data frequency and size.

\section{Conclusion}

This paper presents a comparative analysis of two open-source autonomous driving software platforms—Autoware and Apollo—by examining four core modules as well as their detailed subcomponents. The study further compares the middleware layers, evaluating FastDDS (DDS) and CyberRT in terms of transmission latency, CPU usage, memory consumption, and message loss rate. Results show that while CyberRT achieves lower latency for large-scale data transmission, it also exhibits higher memory usage compared to FastDDS. These findings underscore key trade-offs in both functional scope and middleware performance, thereby providing practical guidance for developers integrating open-source autonomous driving platforms in diverse development environments.

\section*{Acknowledgment}
This work was supported by National Research Foundation of Korea(NRF) grant funeded by the Korea government(MSIT) (No. 2021R1A2C300837014)

\bibliography{main}
\bibliographystyle{plain}

\end{document}